\documentclass[sigconf]{acmart}

\usepackage{subcaption}

\usepackage{xspace}
\makeatletter
\DeclareRobustCommand\onedot{\futurelet\@let@token\@onedot}
\def\@onedot{\ifx\@let@token.\else.\null\fi\xspace}

\def\eg{\emph{e.g}\onedot} 
\def\ie{\emph{i.e}\onedot}

\def\etal{\emph{et~al}\onedot}

\AtBeginDocument{%
  }

\copyrightyear{2024}
\acmYear{2024}
\setcopyright{rightsretained}
\acmConference[CHI EA '24]{Extended Abstracts of the CHI Conference on
Human Factors in Computing Systems}{May 11--16, 2024}{Honolulu, HI, USA}
\acmBooktitle{Extended Abstracts of the CHI Conference on Human Factors in
Computing Systems (CHI EA '24), May 11--16, 2024, Honolulu, HI, USA}
\acmDOI{10.1145/3613905.3650760}
\acmISBN{979-8-4007-0331-7/24/05}

\begin{document}

\makeatletter
\def\@ACM@checkaffil{
    \if@ACM@instpresent\else
    \ClassWarningNoLine{\@classname}{No institution present for an affiliation}%
    \fi
    \if@ACM@citypresent\else
    \ClassWarningNoLine{\@classname}{No city present for an affiliation}%
    \fi
    \if@ACM@countrypresent\else
        \ClassWarningNoLine{\@classname}{No country present for an affiliation}%
    \fi
}
\makeatother

\title{ARtVista: Gateway To Empower Anyone Into Artist}

\author{Trong-Vu Hoang}
\authornote{Equal contribution}
\orcid{0000-0001-7367-1401}
\affiliation{%
  \institution{\textsuperscript{\rm 1} University of Science, VNU-HCM, Ho Chi Minh City, Vietnam}
}
\affiliation{
  \institution{\textsuperscript{\rm 2} Vietnam National University, Ho Chi Minh City, Vietnam}
}

\author{Quang-Binh Nguyen}
\authornotemark[1]
\orcid{0000-0003-1199-3661}
\affiliation{%
  \institution{\textsuperscript{\rm 1} University of Science, VNU-HCM, Ho Chi Minh City, Vietnam}
}
\affiliation{
  \institution{\textsuperscript{\rm 2} Vietnam National University, Ho Chi Minh City, Vietnam}
}

\author{Duy-Nam Ly}
\orcid{0000-0003-4304-2334}
\affiliation{%
  \institution{\textsuperscript{\rm 1} University of Science, VNU-HCM, Ho Chi Minh City, Vietnam}
}
\affiliation{
  \institution{\textsuperscript{\rm 2} Vietnam National University, Ho Chi Minh City, Vietnam}
}

\author{Khanh-Duy Le}
\orcid{0000-0002-8297-5666}
\affiliation{%
  \institution{\textsuperscript{\rm 1} University of Science, VNU-HCM, Ho Chi Minh City, Vietnam}
}
\affiliation{
  \institution{\textsuperscript{\rm 2} Vietnam National University, Ho Chi Minh City, Vietnam}
}

\author{Tam V. Nguyen}
\orcid{0000-0003-0236-7992}
\affiliation{%
  \institution{\textsuperscript{\rm 3} Department of Computer Science \\University of Dayton\\ Ohio, United States}
}

\author{Minh-Triet Tran}
\orcid{0000-0003-3046-3041}
\affiliation{%
  \institution{\textsuperscript{\rm 1} University of Science, VNU-HCM, Ho Chi Minh City, Vietnam}
}
\affiliation{
  \institution{\textsuperscript{\rm 2} Vietnam National University, Ho Chi Minh City, Vietnam}
}

\author{Trung-Nghia Le}
\authornote{Corresponding author. Email address: ltnghia@fit.hcmus.edu.vn}
\orcid{0000-0002-7363-2610}
\affiliation{%
  \institution{\textsuperscript{\rm 1} University of Science, VNU-HCM, Ho Chi Minh City, Vietnam}
}
\affiliation{
  \institution{\textsuperscript{\rm 2} Vietnam National University, Ho Chi Minh City, Vietnam}
}


\begin{abstract}


Drawing is an art that enables people to express their imagination and emotions. However, individuals usually face challenges in drawing, especially when translating conceptual ideas into visually coherent representations and bridging the gap between mental visualization and practical execution. In response, we propose ARtVista - a novel system integrating AR and generative AI technologies. ARtVista not only recommends reference images aligned with users' abstract ideas and generates sketches for users to draw but also goes beyond, 
crafting vibrant paintings in various painting styles. 
ARtVista also offers users an alternative approach to create striking paintings by simulating the paint-by-number concept on reference images, empowering users to create visually stunning artwork devoid of the necessity for advanced drawing skills. We perform a pilot study and reveal positive feedback on its usability, emphasizing its effectiveness in visualizing user ideas and aiding the painting process to achieve stunning pictures without requiring advanced drawing skills.
The source code will be available at \url{https://github.com/htrvu/ARtVista}.
  
\end{abstract}




\begin{teaserfigure}
\includegraphics[width=\textwidth]{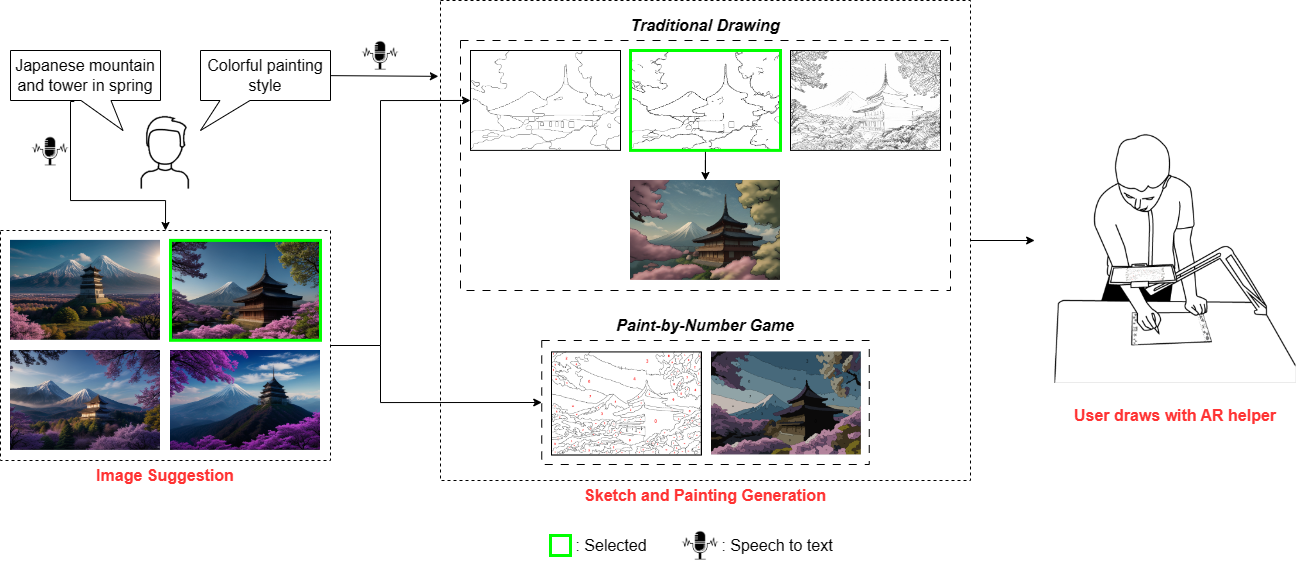}
    \caption{Through AR, the proposed ARtVista system elevates the artistic process, delivering a seamless and immersive experience. Users first input prompts via speech recognition to generate realistic images and then select one to transform it into a specific painting style. After that, the users choose between Traditional Drawing or Paint-by-Number modes to create their preferred painting. Finally, the users draw paintings via an interactive AR interface. 
    }
    \label{fig_teaser}
\end{teaserfigure}
\maketitle

\section{Introduction}

In the quiet strokes of a pencil or the vibrant hues of a digital canvas, drawing has long been a conduit for human imagination and emotions. 
Drawing not only enables us to express moments of inspiration or convey meaningful stories, but also, through rhythmic strokes, has been demonstrated to have a therapeutic effect, aiding in the release of stress and anxiety~\cite{drawreducestress}.

However, achieving the creation of a beautiful picture involves a challenging process.
Firstly, the difficulty lies in translating a conceptual idea into a vivid and coherent visual representation. Even with a preconceived idea in mind, not everyone possesses the ability to envision the complete visual outcome of their drawing, including aspects such as the picture's layout and color. For instance, if a person intends to draw a scene of \textcolor{black}{``a Japanese tower and mountain in spring''}, where should be the placement of the tower and mountain, and what specific details of the surrounding landscape are appropriate? Secondly, the gap between mental visualization and the practical execution of a drawing can be profound, underscoring the importance of artistic skill and proficiency. Not everyone has many whorls on their fingertips, \ie, the capacity to bring their ideas or even reproduce reference images onto paper or any artistic medium. 
Finally, coloring a sketch to create a beautiful picture is not a simple process due to varying levels of individuals' aesthetic feelings. It becomes even trickier when we try to color the sketch in a specific painting style, like watercolor painting. 



To address drawing challenges, we introduce a novel system, namely ARtVista, to facilitate the creation of beautiful drawings for everyone by synthesizing pictures based on ideas of users and then guiding them to draw those pictures through step-by-step sketching and coloring to follow their desired painting style (See Fig.~\ref{fig_teaser}). Users first input their ideas via speech-to-text translation to receive textual prompts. By harnessing the cutting-edge generative AI technology~\cite{rombach2022high}, ARtVista can bring textual descriptions to compelling visuals of remarkable quality and realism. Our system also recommends the users in finding reference images that align with their initial concepts, offering them a comprehensive preview of their upcoming drawing. ARtVista then converts users' selected reference images into intricate sketches by employing precise contour extraction of main objects~\cite{sam, hed}. 
Taking the creative journey further, our system evolves these sketches into vibrant paintings, tailored to users' preferred styles in our pre-defined style list via generative models~\cite{sdedit, controlnet},
resulting in a harmonious and visually stunning output. Finally, inspired by the widely known paint-by-number game, our system offers users an alternative avenue to create impressive paintings based on their ideas through an automatic number painting region process. 
By integrating AR technology, which overlays the composition draft on the drawing paper through a camera video feed, users can unlock their artistic potential and produce visually appealing artwork without needing advanced drawing skills.

We conducted a pilot user study to gather preliminary qualitative insights on the usability of ARtVista. The results of the pilot study yielded the benefits of the proposed system, emphasizing its effectiveness in visualizing users’ ideas and facilitating the painting process. 
Additionally, participants provided valuable feedback highlighting areas for improvement for our system, such as enhancing the drawing experience with AR technology, incorporating a tool to locate the selected color and providing flexibility in adjusting difficulty levels within the Paint-by-Number mode, offering guidance on mixing colors based on the available palette, and supporting additional drawing modes for individuals with varying levels of familiarity or proficiency in drawing. The source code will be available at \url{https://github.com/htrvu/ARtVista}.

In summary, our main contributions are as follows:
\begin{itemize}
    \item We introduced ARtVista, a novel system that incorporates AR technology and generative AI models to enable the creation of beautiful drawings for everyone, without the need for advanced drawing skills. 
    \item We proposed a novel solution that translates users' conceptual ideas into vibrant paintings, adopting a step-by-step guidance process inspired by the paint-by-number game.
    \item We conducted a pilot study to assess ARtVista's usability, showcasing its effectiveness in visualizing user ideas and simplifying the painting process.
\end{itemize}


\section{Related Work}

\subsection{AR Drawing Tool}

In the past, efforts were made to aid the process of drawing and painting using traditional media and tools. Many individuals struggled with positioning and sizing elements on canvas and executing brushstrokes to achieve desired textures. Prior methods~\cite{rivers2012sculpting,flagg2006projector} utilized projectors as guides for painting. While effective, these methods lacked flexibility and lightweight execution, often requiring a variety of media tools. Recently, the development of mobile applications incorporating Augmented Reality (AR) has emerged as a promising tool to enhance physical artistic creation quickly and without the need for specific devices beyond a smartphone. Various works have explored the potential of AR in drawing and painting. Laviole~\etal~\cite{Laviole2012SpatialAR} proposed AR tools that improve drawing speed and ease, blending physical drawing with projected outlines, and even venturing into stereoscopic drawing with its unique challenges and possibilities. Fischer~\etal~\cite{Fischer2006TheAP} showcased a system to enable interactive manipulation of observed scenes and rendering virtual objects seamlessly integrated with the real environment, offering a good experience for both artists and observers. Beyond drawing, Oh~\etal~\cite{Oh2014MobileAR} presented a mobile AR system that visualizes 3D models from design drawings, significantly reducing recognition time and improving accuracy compared to traditional methods. Ryffel~\etal~\cite{Ryffel2017AMA} demonstrated an interactive application for painting recoloring, achieving realistic color edits with layered alpha channels and an automatic segmentation algorithm for efficient deployment. However, a notable limitation of these early works lies in their dependence on pre-existing objects and outlines, potentially constraining the user's creative freedom and limiting the expressive range and originality of the final artwork. Meanwhile, our ARtVista system assists users in materializing diverse imagery from their conceptual musings via generative AI.

\subsection{Image Generation}


Image generation has evolved through diverse paradigms, including Generative Adversarial Networks~\cite{sauer2023stylegan, liao2022text, kang2023scaling}, autoregressive models~\cite{kingma2019introduction, lee2022autoregressive}, and diffusion models~\cite{rombach2022high, saharia2022photorealistic, balaji2022ediffi, nichol2021glide}. Notably, diffusion models have pushed the boundaries of image generation, showcasing a remarkable ability to synthesize high-fidelity images guided by text. To enhance control in generating output of diffusion models, ControlNet~\cite{controlnet} has emerged as an approach for generation based on specified conditions, such as canny edge, sketching, depth, and more. However, advancements in diffusion models have presented challenges in balancing generation efficiency and effectiveness, as these models often require numerous diffusion steps to produce high-quality images, making it impractical for real-time applications. To address this limitation, we employed LCM-LoRA~\cite{lcmlora} in our proposed ARtVista system to accelerate the generation speed for practical application.


\subsection{Image to Sketch}

Traditional image-to-sketch methods have relied heavily on classic computer vision techniques, particularly edge detection algorithms~\cite{canny1986computational,hed}. These methods analyze image information to identify significant edges and translate them into lines, often resulting in sharp, detailed sketches. However, they can struggle with complex textures and lighting, leading to overly simplistic or noisy outputs. The advent of neural style transfer~\cite{gatys2015neural} offered a more holistic approach, allowing the transfer of a desired artistic style, including sketching styles, onto an image. Early methods ~\cite{gatys2016image, ulyanov2016texture} relied on convolutional neural networks to capture style features and apply them to the content image. More recent work has seen the integration of generative adversarial networks (GANs)~\cite{yi2020line, liu2021deep, isola2017image} into style transfer frameworks, further enhancing the quality of the generated sketches. Through adversarial training, GANs learn to capture the artistic essence of sketch styles and apply them to input images, often producing more artistic and expressive sketches than previous methods.

\section{ARtVista - Multimodal AR Generative Drawing System}

\subsection{Conceptualization}
\label{sec:conceptualization}

In light of the aforementioned challenges of drawing that users encounter, we introduce ARtVista, a portable system aimed at facilitating the creation of beautiful pictures by users. We conceptualize ARtVista as a mobile application designed to aid users in easily creating their artwork wherever they may be.

\textbf{ARtVista Interface} is an AR application that leverages a user interface for generating reference pictures and supports users in the painting process. Fueled by the potent capabilities of generative AI, ARtVista empowers users to materialize their imaginative concepts into  tangible artistic expressions by providing a user-friendly prompt entry system. To enhance the user experience and streamline the input process, ARtVista incorporates an advanced speech recognition feature, the app incorporates a speech recognition feature. This provides users with the ability to input queries hands-free, addressing the need for hands-on engagement in drawing activities where users may be occupied with other tasks. Verbal input-based prompt entering also aligns with the proposed physical setup of ARtVista, as mentioned in the following part. 

\begin{figure*}[t!]
    \centering
\includegraphics[width=\textwidth]{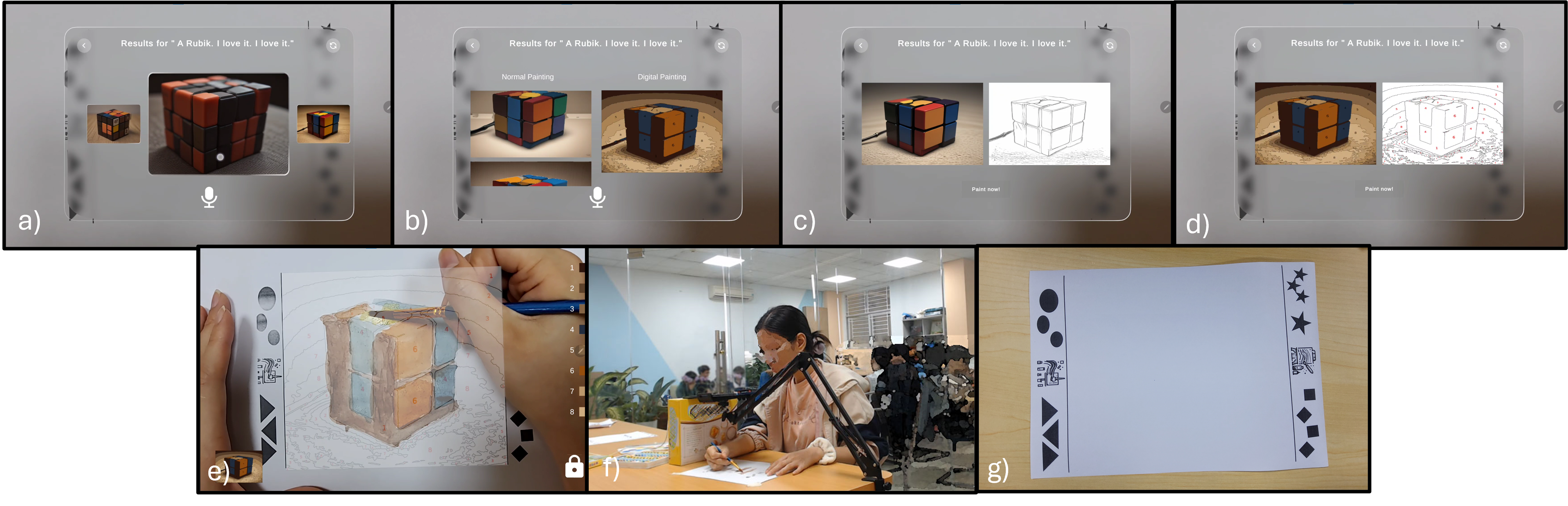}
     \caption{ARtVista interface. a) Realistic reference image suggestion. b) Painting-style picture generating. c)-d) Composition of painting and sketch. e) AR view. f) Physical setup for ARtVista. g) Designed drawing paper.}
    \label{fig:ar_interface}
\end{figure*}

ARtVista's generative AI module allows users to craft lifelike pictures through prompt-driven interactions. This stems from the need to artistically create the scenes or things of a number of users instead of capturing them using a camera. In addition, ARtVista provides users with the capability to convert the generated realistic-style pictures into a painting-style format, enhancing the ease and convenience of the subsequent painting process.

However, painting from a reference picture is a big challenge for users who have poor drawing skills. This issue leads them to spend more time calculating the proportion radio in the drawing. To help users address this problem, ARtVista incorporate a pass-through AR feature which augments the composition sketch of the painting-style picture they selected before. Thanks to this feature, they can easily complete their artwork with less effort compared to traditional drawing from the reference method. While ARtVista can function with AR/MR headsets, our primary focus is on the integration of the application for widely utilized personal devices like smartphones and tablets. When users engage with the app, the drawing interaction involves looking through the screen of their tablet or mobile device to execute the painting process.

\textbf{Drawing setup}: 
In addition to utilizing ARtVista for painting, we strongly recommend users enhance their creative workspace by incorporating a desk-mount arm to elevate the tablet. 
This helps users stabilize the tablet, making it easier for viewing the screen while drawing. Thus, utilizing this setup, users employing the voice recognition feature will find it more convenient and quicker than entering prompts manually through the keyboard.

\textbf{Interaction Flow} of painting a creative image using ARtVista is as follow (Fig. \ref{fig_teaser}). Users first input the prompt to get realistic images generated by ARtVista using the speech recognition function. After that, users choose one of these pictures to transform it into a specific painting style. Next, users select the mode to start the drawing process. Currently, ARtVista supports two modes for users to create their paintings, which are normal painting and painting by number. Normal painting mode assists users in drawing with the augmentation of the composition draft via a camera video feed with three levels of difficulty. Meanwhile, the Paint-by-Number mode displays a paint-by-number template and a color palette on the UI, users can fill in colors by matching them with corresponding boxes in the template. 

\begin{figure*}[t!]
    \centering
    \includegraphics[width=\linewidth]{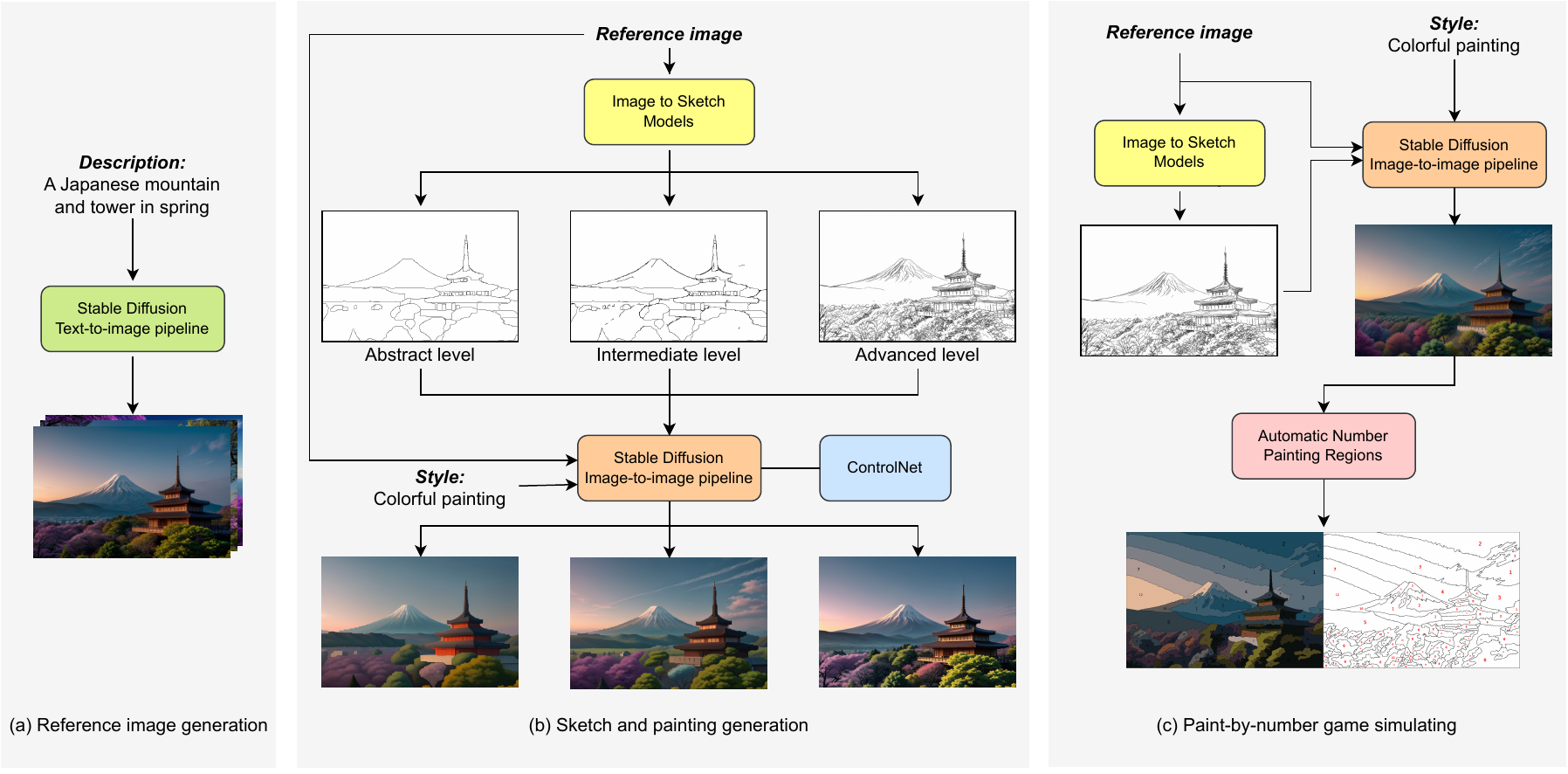}
     \caption{Overview of image generation workflow with three main capabilities: reference image generation, sketch and painting generation, and paint-by-number game simulating.}
    \label{fig:image_gen_framework}
\end{figure*}


\subsection{Prototype Implementation}



\subsubsection{AR Interface: }

For the development of the ARtVista interface, we utilize Unity as the primary platform. This interface establishes the communication with Image Generation Framework (Part. ~\ref{ImageGenerationFramework}), which was pre-deployed in our local server, via HTTP APIs to use its functionality. Additionally, to enable users to express their creative input through natural speech interactions, we incorporate OpenAI's Whisper~\cite{whisper} model provided by the Hugging Face service for recognizing and transforming the user's speech into textual input. The language we use in our system is English. The illustration for ARtVista's interactive AR interface with the interaction flow mentioned in Sec.~\ref{sec:conceptualization} is depicted in Fig.~\ref{fig:ar_interface}.

After users select the desired painting-style picture for drawing, ARtVista overlays the composition sketch onto the live video feed captured from the device's camera. To facilitate users in drawing pictures seamlessly without being influenced by the shifting of the drawing paper with augmented information, we design a special texture for the drawing paper, as seen in Fig.~\ref{fig:ar_interface}-g, and employ Vuforia to precisely track that paper location.
By effectively mitigating the impact of paper movement, users can concentrate on their artistic expressions without spending much time aligning the unfinished painting to match the augmented sketch.

\subsubsection{Image Generation Framework: }
\label{ImageGenerationFramework}
To empower users to translate their mental concepts into tangible artistic expressions regardless of their artistic skill, we design a comprehensive image generation framework as seen in Fig.~\ref{fig:image_gen_framework}. This framework takes user prompts to generate reference lifelike images, transforming selected images into various sketch versions corresponding with desired painting style pictures. This process aids users in effortlessly obtaining drafts for their paintings, assisting in shaping composition, especially for those unfamiliar with proportional drawing.


Our framework utilizes Stable Diffusion~\cite{rombach2022high} with the Realistic Vision checkpoint, a cutting-edge text-to-image model, to produce reference real images based on the description of users about the thing they want to draw. 
Additionally, we employ techniques like CPU offloading and attention layer acceleration~\cite{xFormers} to optimize the memory consumption of the generation process. 
By integrating LCM-LoRA~\cite{lcmlora}, we further enhance the generation speed, enabling us to present users with numerous choices within a suitable timeframe.



\begin{figure*}[t!]
    \centering
    \includegraphics[width=\linewidth]{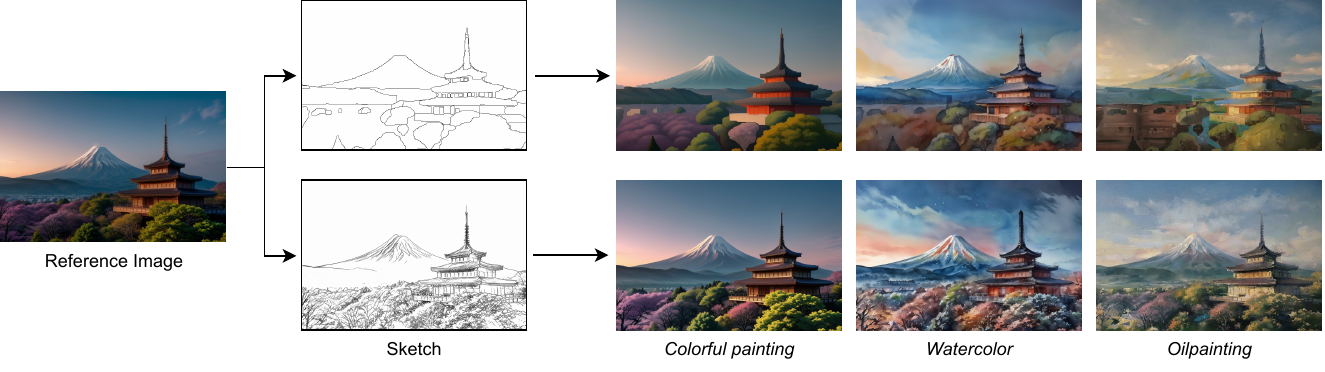}
     \caption{The illustration for the output of some painting styles of our framework: Colorful painting, watercolor, and oil painting.}
    \label{fig:painting_styles}
\end{figure*}

Deep learning models are employed to generate sketch versions of those images. 
We offer users three distinct sketching levels that progressively enhance in detail, as seen in Fig.~\ref{fig:image_gen_framework}-b. 
\textit{Abstract-level sketches} are obtained by leveraging SAM model~\cite{sam} to semantically segment the given image, followed by region mining, resulting in visually engaging abstract sketch suitable for beginners and children. \textit{Intermediate-level sketches}, a baseline for everyone to start with our system, are created by fusing edge detection algorithms~\cite{hed, sam} to enhance the sketch's detail and intricacy. \textit{Advanced-level sketches} are achieved through Line Art style transfer model~\footnote{\url{https://huggingface.co/spaces/awacke1/Image-to-Line-Drawings}} for capturing intricate details and nuances, providing users with advanced sketching options. 
Stable Diffusion~\cite{rombach2022high} with DreamShaperV8 checkpoint, which is known for the proficiency in generating artistic images, is employed to transform users' sketches into colored paintings according to their preferred painting style. ControlNet~\cite{controlnet} and LCM-LoRA~\cite{lcmlora} also are integrated to guarantee that the resulting painted images align with the strokes of the original sketches. The output of different painting styles produced by our framework is illustrated in Fig.~\ref{fig:painting_styles}.

\begin{figure*}[t!]
    \centering
    \includegraphics[width=\linewidth]{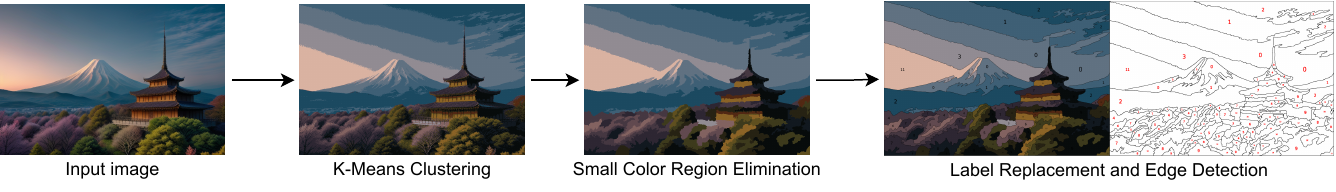}
     \caption{The workflow of automatic number painting regions process.}
    \label{fig:paint_by_number_process}
\end{figure*}

To create the template for the Paint-by-Number mode of ARtVista, we perform an \textit{automatic number painting regions} procedure. This involves utilizing the K-Means algorithm~\cite{kmeans} along with a graph traversal algorithm to compress and cluster color bits. Subsequently, we extract edge details to outline the boundaries of each region and allocate the cluster's index to these delineated areas.
Figure~\ref{fig:paint_by_number_process} illustrates the workflow of this process.



\section{Pilot study}

\begin{figure*}[t!]
    \centering
    \includegraphics[width=\linewidth]{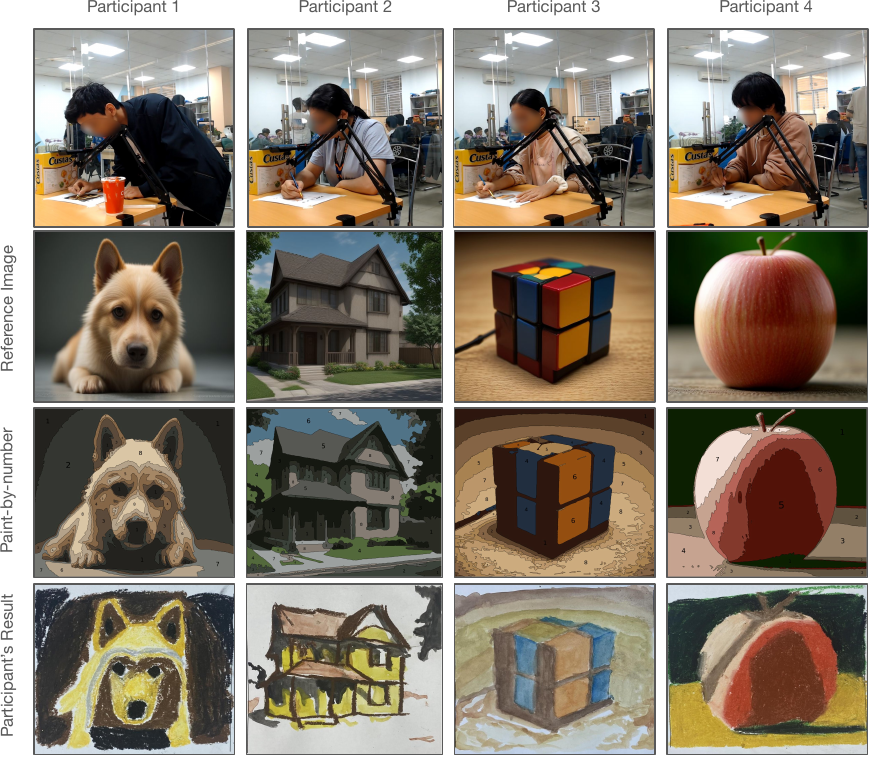}
     \caption{Paintings created by participants using ARtVista.}
    \label{fig:participants_paintings}
\end{figure*}


In our preliminary investigation, we conducted a pilot study aimed at obtaining initial feedback on ARtVista. 
The primary objective was to explore the efficacy of ARtVista in facilitating users to transform their mental concepts into tangible artistic expressions and offer assistance throughout the drawing process, as well as shed light on potential improvements for our system.

\subsection{Study Setup}
We invited four participants to experience our system. All participants were students at our faculty with ages ranged between 19 and 20 and non-native English speaker. Among them, two girls exhibit basic painting skills, whereas the two boys rarely engage in drawing activities.

In this study, all participants were equipped with a Samsung Tab S7 device for the execution of ARtVista. Additionally, we also supplied wax color crayons, watercolors, and pre-designed drawing paper to enable users to engage in painting while utilizing the support of the app. To optimize the viewing experience of the augmented reality (AR) screen during drawing sessions, a desk-mount arm was employed to elevate the tablet, aligning its screen orientation parallel to the table surface


The assigned task for participants involved utilizing ARPainting to create desired images and employing the Painting-by-number mode for guidance. Due to the app's language configuration, participants are required to use English while engaging with the application. Throughout the process, we closely monitored user behaviors, conducting post-painting interviews immediately after each participant completed their artwork. The drawing session for each participant averaged around 30 minutes, followed by a 10-minute post-interview period. The whole study was video-recorded for our analysis purposes.

\subsection{Initial Insights}

During the observation of users interacting with ARtVista, a primary challenge emerged in their ability to generate images based on verbal input. This difficulty predominantly stemmed from issues related to the performance of the speech recognition feature and variations in users' pronunciation. As a temporary solution during the study, participants were provided with an additional smartphone equipped with the Google Translate application's voice feature to aid in articulation. However, this was acknowledged as a provisional measure for study completion.

In the Paint-by-Number mode, P1 expressed difficulty with the initial 16-color setup, finding it time-consuming. In response, we adjusted the IGF module to require only 8 colors, aiming to streamline the drawing session. P1 also suggested a customization feature, allowing users to choose their preferred number of colors, thus personalizing the difficulty level of this mode.

Meanwhile, P4 tried to tap on a color-1 button in the color palette panel to expect the system to highlight the areas he needed to fill color-1 but did not work. P4 also conveyed the difficulty he faced in locating the colors recommended by the app and keeping track of the available colors. This is evident in his result painting, 
where the color tone did not truly align with the output of the Paint-by-Number mode (Fig. \ref{fig:participants_paintings}). 
He wished for a mechanism within the system that would allow him to register existing colors and provide guidance to mix a new color to enhance his painting experience.

Additionally, during the time experienced our system, P2 recommended that it would be more engaging if our system incorporated a paint-by-layer mode, tailored for individuals with some familiarity with watercolor painting or oil painting, as it aligns with their preferred learning approach for those styles and they frequently struggle with blending colors accurately. P2 also suggested a support for mimicking the artistic styles of popular artists like Van Gogh and Leonardo da Vinci, specifically catering to individuals with proficient drawing techniques.

In terms of the drawing experience, users expressed a slight discomfort when drawing through the screen compared to the tactile sensation of traditional drawing. Notably, participants occasionally diverted their gaze from the tablet screen, opting to look directly at the paper while drawing. This behavior was attributed to the perceived ease of drawing without augmented reality (AR) support. Participants mentioned concerns such as color smudging, particularly noticeable when using wax color crayons,
and the difficulty in estimating the distance between the pen nibs and the drawing paper. 
These issues may result in unexpected strokes and hinder the pace of their painting process.

\section{Conclusion and Future Work}

In this paper, we introduced ARtVista, a novel system that enables the creation of beautiful drawings for everyone regardless of advanced drawing skills.
By incorporating cutting-edge generative AI models and AR technology, ARtVista can synthesize images based on users' conceptual ideas and guide them through step-by-step sketching and coloring processes to achieve visually captivating artworks. 
We conducted a pilot study to explore the effectiveness of ARtVista in supporting users to visualize and picture their ideas. 

Based on the feedback gathered in this study, we will tackle the existing issues in the current prototype and enhance both the user experience and the variety of drawing modes in the future. 
Firstly, we intend to improve the drawing experience through the screen of the device by providing more accurate color visualization on the screen and integrating the depth estimation feature of the device's camera to simulate the interaction between the pen and the drawing paper.
Next, we will consider the possibility of specific painting styles (\eg Impressionist and Baroque styles) and drawing modes tailored for individuals with some familiarity with drawing skills. To replicate popular artistic painting styles more effectively, we intend to utilize more robust checkpoints for the Stable Diffusion model. The upcoming drawing mode that we plan is paint-by-layer, which can be achieved by integrating deep learning models capable of stroke prediction, like PaintTransformer~\cite{painttransformer}.

\section*{Acknowledgments}

This research is funded by Vietnam National Foundation for Science and Technology Development (NAFOSTED) under Grant Number 102.05-2023.31. 

Additionally, Dr. Tam V. Nguyen is supported by National Science Foundation (NSF) under Grant Number 2025234.

\bibliographystyle{ACM-Reference-Format}
\bibliography{sample-base}

\end{document}